%% file: main.tex
\documentclass[10pt,twocolumn,letterpaper]{article}

\usepackage[pagenumbers]{iccv} 

\input{preamble}

\definecolor{iccvblue}{rgb}{0.21,0.49,0.74}
\usepackage[pagebackref,breaklinks,colorlinks,allcolors=iccvblue]{hyperref}

\def\paperID{7247} 
\def\confName{ICCV}
\def\confYear{2025}

\title{RoboMatrix: A Skill-centric Hierarchical Framework for Scalable \\ Robot Task Planning and Execution in Open-World}

\author{
Weixin Mao$^{1}$\thanks{Equal contribution.} \thanks{Project leader}, \quad  Weiheng Zhong$^{2*}$, \quad Zhou Jiang$^2$, \quad Dong Fang$^3$, \quad Zhongyue Zhang$^2$, \\
\quad Zihan Lan$^4$, \quad Haosheng Li$^5$, \quad Fan Jia$^4$,
Tiancai Wang$^4$, \quad Haoqiang Fan$^4$,\quad Osamu Yoshie$^{1}\thanks{Corresponding author.}$\\
{\tt\small$^1$Waseda University}
{\tt\small$^2$Beijing Institute of Technology} \\
{\tt\small$^3$The Chinese University of Hong Kong} \\
{\tt\small$^4$MEGVII Technology $^5$Chinese Academy of Sciences}
}

\usepackage[linesnumbered, ruled]{algorithm2e}
\usepackage{colortbl}
\usepackage{epigraph}
\usepackage{multirow}
\usepackage{amssymb} 

\usepackage{multirow}
\usepackage{graphicx}
\usepackage{tabularx}
\usepackage{diagbox}

\begin{document}
\maketitle
\input{sec/0_abstract}    
\input{sec/1_intro}
\input{sec/2_related}

\input{sec/3_methods}

\input{sec/4_exp}
\input{sec/5_conclusion}

\clearpage
{
    \small
    \bibliographystyle{ieeetr}
    \bibliography{main}
}

\input{sec/X_suppl}

\end{document}

%% file: preamble.tex
\usepackage[dvipsnames]{xcolor}


%% file: sec/0_abstract.tex
\begin{abstract}
Existing robot policies predominantly adopt the task-centric approach, requiring end-to-end task data collection. This results in limited generalization to new tasks and difficulties in pinpointing errors within long-horizon, multi-stage tasks. To address this, we propose \textbf{RoboMatrix}, a skill-centric hierarchical framework designed for scalable robot task planning and execution in open-world environments. RoboMatrix extracts general meta-skills from diverse complex tasks, enabling the completion of unseen tasks through skill composition. Its architecture consists of a high-level scheduling layer that utilizes large language models (LLMs) for task decomposition, an intermediate skill layer housing meta-skill models, and a low-level hardware layer for robot control.
A key innovation of our work is the introduction of the first unified vision-language-action (VLA) model capable of seamlessly integrating both \textbf{movement} and \textbf{manipulation} within one model. This is achieved by combining vision and language prompts to generate discrete actions.
Experimental results demonstrate that RoboMatrix achieves a 50\% higher success rate than task-centric baselines when applied to unseen objects, scenes, and tasks.
To advance open-world robotics research, we will open-source code, hardware designs, model weights, and datasets at \href{https://github.com/WayneMao/RoboMatrix}{https://github.com/WayneMao/RoboMatrix}.


\end{abstract}


%% file: sec/1_intro.tex
\section{Introduction}
\label{sec:intro}

\setlength{\epigraphwidth}{1.0\columnwidth}
\renewcommand{\epigraphflush}{center}
\renewcommand{\textflush}{flushepinormal}
\renewcommand{\epigraphsize}{\footnotesize}

\epigraph{\textcolor{black}{``The more things change, the more they stay the same.''}}
{\textcolor{black}{\textit{Jean-Baptiste Alphonse Karr, French writer, 1984}}}



\begin{figure}[!t]
  \centering
   \includegraphics[width=1.0\linewidth]{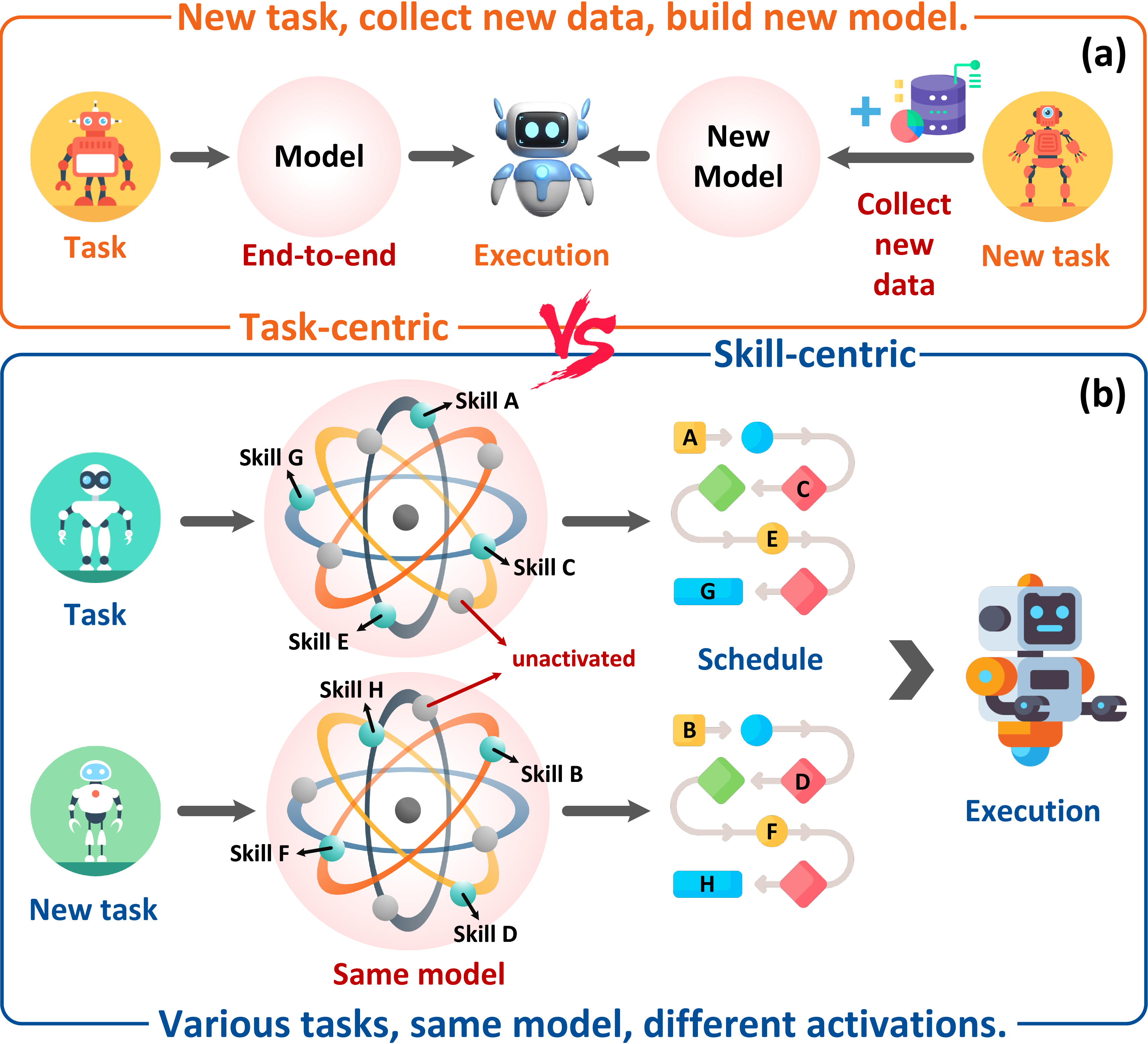}
   \caption{\textbf{Task-Centric vs. Skill-Centric.} (a) The task-centric paradigm requires collecting new data and training a new model for each new task. (b) The skill-centric paradigm enables zero-error task generalization by activating different skill responses within one fully trained VLA skill model.}
   \label{fig:skill_centric}
   \vspace{-10pt}
\end{figure}

Recent advancements in vision-language models (VLMs)~\cite{llava1,llava2,wang2024qwen2vl,DreamLLM} have enabled novel vision-language-action (VLA) frameworks~\cite{pi0,GR1,OpenVLA,octo_2023,RT2,adriver-i,liu2024rdt} that integrate visual perception with language-guided action prediction. These end-to-end approaches demonstrate promising results in manipulation tasks, yet their \textit{task-centric} nature---as illustrated in Fig.~\ref{fig:skill_centric}(a)---imposes fundamental limitations for open-world scenarios. Specifically: (1) Complete task demonstration requirements lead to exponential data growth with task complexity~\cite{3DDP,lin2024data}; (2) The end-to-end architectures struggle with novel task compositions~\cite{li2024manipllm}; (3) Black-box learning mechanisms hinder error diagnosis~\cite{honerkamp2023n}. 
These limitations stem from conflating three core robotic capabilities: environment \textit{perception}, sub-task \textit{reasoning}, and physical \textit{interaction}---capabilities that traditional methods address in isolation through imitation learning (IL)~\cite{ACT,chi2023DiffusionPolicy,3DDP,Mobilealoha} or reinforcement learning (RL)~\cite{xia2021relmogen,honerkamp2023n,sun2022fully}, but fail to generalize synergistically.

The fundamental challenge for open-world manipulation lies in generalization to unseen scenarios and tasks, which requires but is rarely achieved by existing methods in recomposing meta-skills for novel task specifications. Existing solutions bifurcate into two limited paradigms:
(1) \textit{Task-specific traditional methods} (IL/RL~\cite{Mobilealoha,sun2022fully}) that tightly couple perception-action spaces, suffering catastrophic failures with novel object-task pairings;
(2) \textit{LLM-Based methods}~\cite{pi0,GR1,octo_2023,OpenVLA} that despite leveraging arge language models (LLMs) priors, inherit the task-centric pitfalls: prohibitive demonstration costs, limited skill transfer, and undiagnosable errors.
Our key insight is that \textit{decoupling skill learning from task composition} enables: (a) Meta-skill reuse across tasks, (b) Transparent error diagnosis, and (c) Data-efficient adaptation---preserving foundation models' strengths while overcoming their architectural constraints.

To overcome these limitations, we propose RoboMatrix --- a skill-centric hierarchical framework that enables \textit{compositional task execution} through meta-skill recombination. 
As shown in Fig.~\ref{fig:skill_centric}(b), our fully-trained VLA skill model enables zero-shot task generalization by \textbf{dynamically activating specific skill responses} (e.g., "Grasp-Response," "Move-Response") based on environmental observations and task context. This paradigm facilitates new task completion via skill recombination, eliminating the need for additional data collection or model fine-tuning.
To achieve task decomposition and arrange the skills for new tasks, RoboMatrix adopts a hierarchical framework. It is structured into three layers: a scheduling layer, a skill layer, and a hardware layer. The scheduling layer employs a general LLM to decompose the task and select appropriate skill models. The skill layer comprises the meta-skill models. The hardware layer includes the physical robot and a communication system, which facilitates seamless integration with higher-level modules.

Compared to the task-centric paradigm, RoboMatrix's skill-centric approach significantly improves interpretability, data efficiency, and generalization. 
Specifically, as shown in Tab.~\ref{tab:long_horizon}, the skill-centric paradigm achieves a 40\% higher success rate on hard-level tasks. More importantly, in \textit{Level V} generalization scenarios (Tab.~\ref{tab:level}), our method outperforms task-centric baselines by 50\% in success rate, validating its superior capability in handling novel task compositions and environmental variations.
In summary, our contributions are:
\begin{itemize}
\item A skill-centric, hierarchical framework for scalable robot task planning and execution in open-world environments.
\item A novel unified VLA model that integrates vision and language prompts to generate both movement and manipulation actions, enhancing coordination in complex tasks.
\item Demonstrated superior generalization to novel objects, scenes, and tasks, achieving a 50\% higher success rate than the task-centric baseline.
\end{itemize}



%% file: sec/2_related.tex
\section{Related Works}
\label{sec:}



\paragraph{Task Planning.}
Addressing long-horizon tasks has long been a central focus in robotics research \cite{d2024achieving}. Behavior trees have been extensively applied for state switching within a finite set of tasks \cite{Behavior, unified_behavior}. However, their effectiveness is constrained by fixed control flows, making them less adaptable to dynamic environments. \cite{Multi-stage} leverages neural networks for high-level subtask selection to handle complex and variable tasks but these approaches still face challenges when dealing with tasks that require reasoning in open-world scenarios.

With the rapid advancement of LLMs, it has become feasible to tackle long-horizon complex tasks in open-world environments.
Numerous studies have employed LLMs as high-level task planners, translating language instructions into executable subtasks for robots \cite{saycan2022arxiv, planner, singh2023progprompt,hu2024look, huang2023grounded}. Some resarech utilizes LLMs to decompose tasks and generate code for accomplishing sub-tasks \cite{mei2024replanvlm, liang2023code}.
Furthermore, numerous studies incorporate multimodal foundational models that leverage scene understanding and language reasoning capabilities to address long-horizon complex tasks 
 \cite{duan2024manipulate, jiang2022vima, VoxPoser}.




\vspace{-10pt}
\paragraph{Task-centric and skill-centric.}
Task-centric approaches aim to enhance the performance of specific tasks, often necessitating the collection of task-specific data or the design of specialized methods \cite{fu2024mobile, chi2024universal, RT1, iyer2024open}. This process is typically time-consuming and labor-intensive, posing challenges in generalizing these methods to other tasks.
On the other hand, leveraging the high-level task planning capabilities of LLMs allows for the definition of multiple subtasks to accomplish various complex tasks \cite{ding2023task, huang2022language,wang2024describe}. Nonetheless, each subtask requires specific data or methods for implementation. and when a task falls outside the predefined set, the overall execution may fail.

In contrast, skill-centric approaches emphasize the development of generalizable skills that can be reused across different tasks \cite{shahvalue, hangl2017autonomous}. By composing various meta-skills, it is possible to flexibly accomplish a wide range of tasks without the need for task-specific data collection or redesign.
In this paper, we focus on acquiring meta-skills and building a skill database to enable the completion of diverse tasks.



\begin{figure*}[htbp] 
  \centering
   \includegraphics[width=0.95\linewidth]{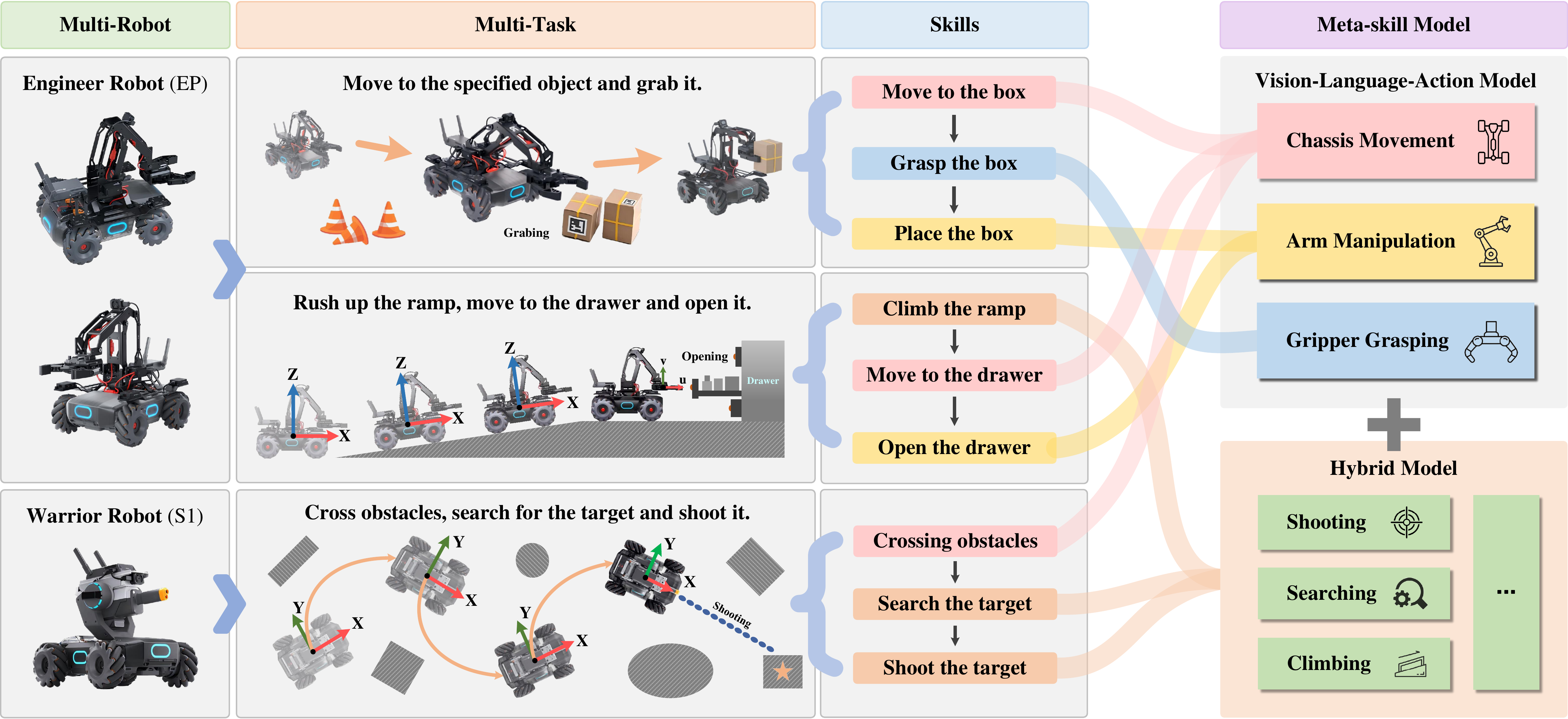}
   
   \caption{\textbf{Inspiration of the skill-centric method.}
    Robots with different modalities can perform different tasks, and robots with the same modality can be used in various scenarios. We extract similar elements from the multitude of diverse robotic tasks, defining these elements as meta-skills and storing them in a skill list. Then, these skills are used to train the Vision-Language-Action (VLA) model or to construct hybrid models, which can eventually lead to a skill model capable of adapting to new tasks.
   }
   
   \label{fig:meta-skills}
   \vspace{-5pt}
\end{figure*}

\vspace{-5pt}
\paragraph{LLM-driven research in Embodied AI.}
Recent advancements in large language models have demonstrated promising results in embodied intelligence. ~\cite{VoxPoser, ReKep, mei2024replanvlm, wang2024solving, qian2024thinkgrasp, mu2024embodiedgpt, jiang2022vima, chatgpt2robotics} directly utilize ChatGPT ~\cite{GPT1,GPT2, GPT4} to construct agents for task decomposition and planning.
Multimodal large models, such as~\cite{li2024manipllm, PaLM-E, chen2024spatialvlm, GR1}, integrate visual, language, and other modal information to enhance robots' understanding and interaction with the environment. These models harness the power of pre-training on large-scale datasets and fine-tuning with task-specific data to achieve state-of-the-art performance in various embodied AI tasks. 
On the other hand, Vision-Language-Action (VLA) models, exemplified by ~\cite{RT2, RT-H, OpenVLA, wen2024tinyvla, RoboFlamingo, adriver-i, xu2024drivegpt4, preACT}, take a step further by directly combining visual and language information with robot action decision-making. 


%% file: sec/3_methods.tex
\begin{figure}[!b]
  \centering
  \includegraphics[width=0.9\linewidth]{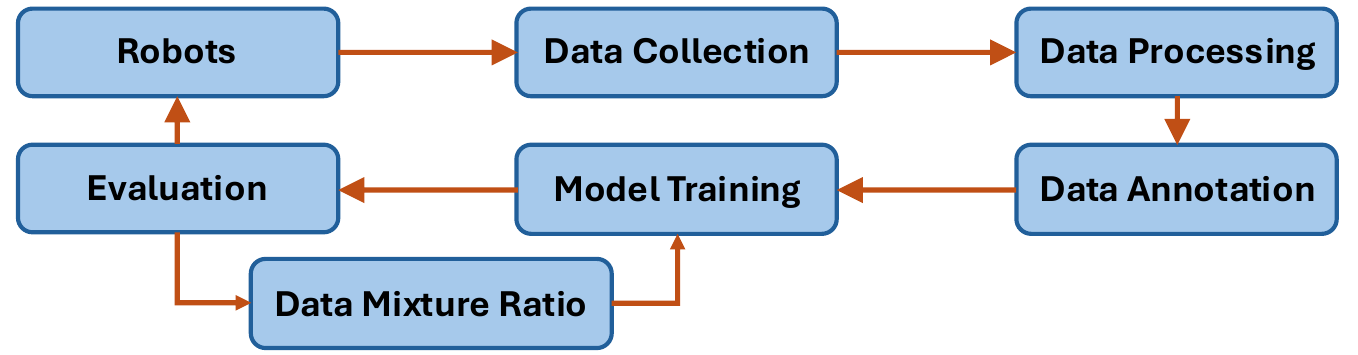}
   \caption{\textbf{The pipeline of data engine.}}
   \label{fig:data_engine}
   \vspace{-0.1in}
\end{figure}

\begin{figure*}[htbp] 
  \centering
   \includegraphics[width=0.95\linewidth]{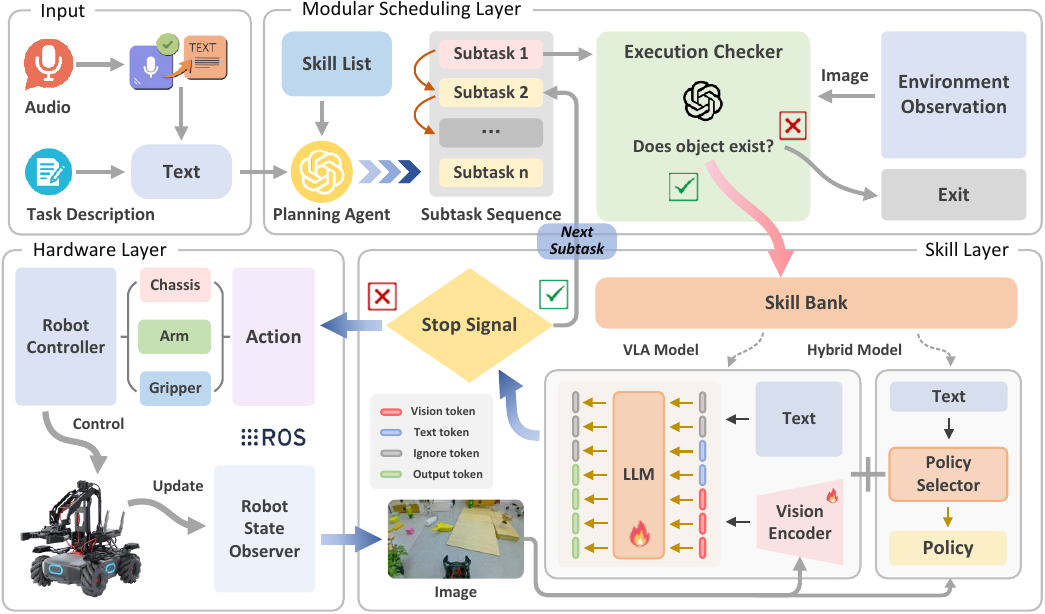}
   
   \caption{\textbf{RoboMatrix Overview.}
    The system accepts the task description in either text or audio format. The text can be entered manually, while the audio is converted into text format by the audio-to-text module.
    The \textbf{Modular Scheduling Layer} serves as the high-level planner of the system. The agent decomposes complex tasks into an ordered sequence of subtasks based on the robot's skill list and adds them sequentially to the execution queue.
    Before executing a subtask, the execution checker verifies its executability by determining whether the object to be manipulated or grasped is present in the scene based on the robot's environment observations.
    The \textbf{Skill Layer} maps the description of subtasks to robot actions using either the hybrid model or the VLA model, with the action including a stop signal to determine whether the current subtask is complete.
    The \textbf{Hardware Layer} manages the controller and stage observer of the robot, with the controller converting actions into control signals and the stage observer continuously updating the robot's state and image in real-time.}
   
   \label{fig:framework}
   \vspace{-10pt}
\end{figure*}

\section{Methods}

For scalable task planning and execution in open-world environments, we propose RoboMatrix, a hierarchical framework built on a skill-centric paradigm.
We first discuss the skill-centric pipeline: how to construct a set of meta-skills for complex tasks and a unified skill database (see Sec.~\ref{sec:skill-centric-ppl}).
Based on these predefined skills, we detail our novel skill models, including the vision-language-action and hybrid models (see Sec.~\ref{sec:skill-models}).
Then we elaborate the operational mechanism of RoboMatrix: how this framework works on real-world robots (see Sec.~\ref{sec:robomatrix-framework}).



\subsection{Skill-centric Pipeline}
\label{sec:skill-centric-ppl}
Intuitively, robots can perform a theoretically infinite variety of tasks in the open world, but it is resource-intensive and time-consuming to collect every task-specific data whenever a new task is established. Therefore, a natural question arises: Are there invariant elements that exist among different tasks?

\vspace{-10pt}
\paragraph{Meta-skills.}

In fact, similar to atoms, a complex task consists of a finite and enumerable set of indivisible minimum meta-skills, which is the core inspiration of the skill-centric method. As illustrated in Fig.~\ref{fig:meta-skills}, despite the diversity of robotic tasks, a commonality emerges in primitive hardware units (e.g., mobile chassises, robotic arms) and their interaction patterns with the environment (e.g., movement, manipulation), which serve to define the meta-skills of the robot. 
For instance, the mobile chassis can achieve the functionality of movement in the open-world environment. In different complex tasks, this function may be utilized in specific processes such as ``move to the box'', ``move to the drawer'', or ``crossing obstacles'' and by any robot equipped with a mobile chassis. Due to the similarity of the ``move to'' action and the uniqueness of the ``crossing'' action, we can define ``move to $<$object$>$" and ``crossing $<$obstacles$>$" as two meta-skills, which are not limited to a single task or a single robot.
For other primitive hardware units, the same strategy can be employed to extract meta-skills.







\vspace{-10pt}
\paragraph{Skill Database and Data Engine.}
The construction of the Skill Database is divided into two distinct phases: the \textbf{\textit{cold-start phase}} and the \textbf{\textit{scaling-up phase}}. During the cold-start phase, we collect diverse complete task data and, based on the previously mentioned meta-skill division rules, partition this data into skill-specific data clips. In the scaling-up phase, we employ a skill-centric methodology to collect skill data directly, significantly expanding the dataset by increasing the quantity and diversity of skill data.

Furthermore, we develop an efficient data engine to enhance the iterative retraining process, as illustrated in Fig.~\ref{fig:data_engine}. Our trained model is first deployed and tested on physical robots. After evaluation, we collect additional skill data or adjust the proportion of different skill data in the dataset (i.e., the data mixture ratio) to refine underperforming skills while maintaining balance. The model is then retrained with the updated dataset to enhance performance in terms of task completion accuracy and generalization capability.




\begin{figure}[!t]
  \centering
   \includegraphics[width=0.9\linewidth]{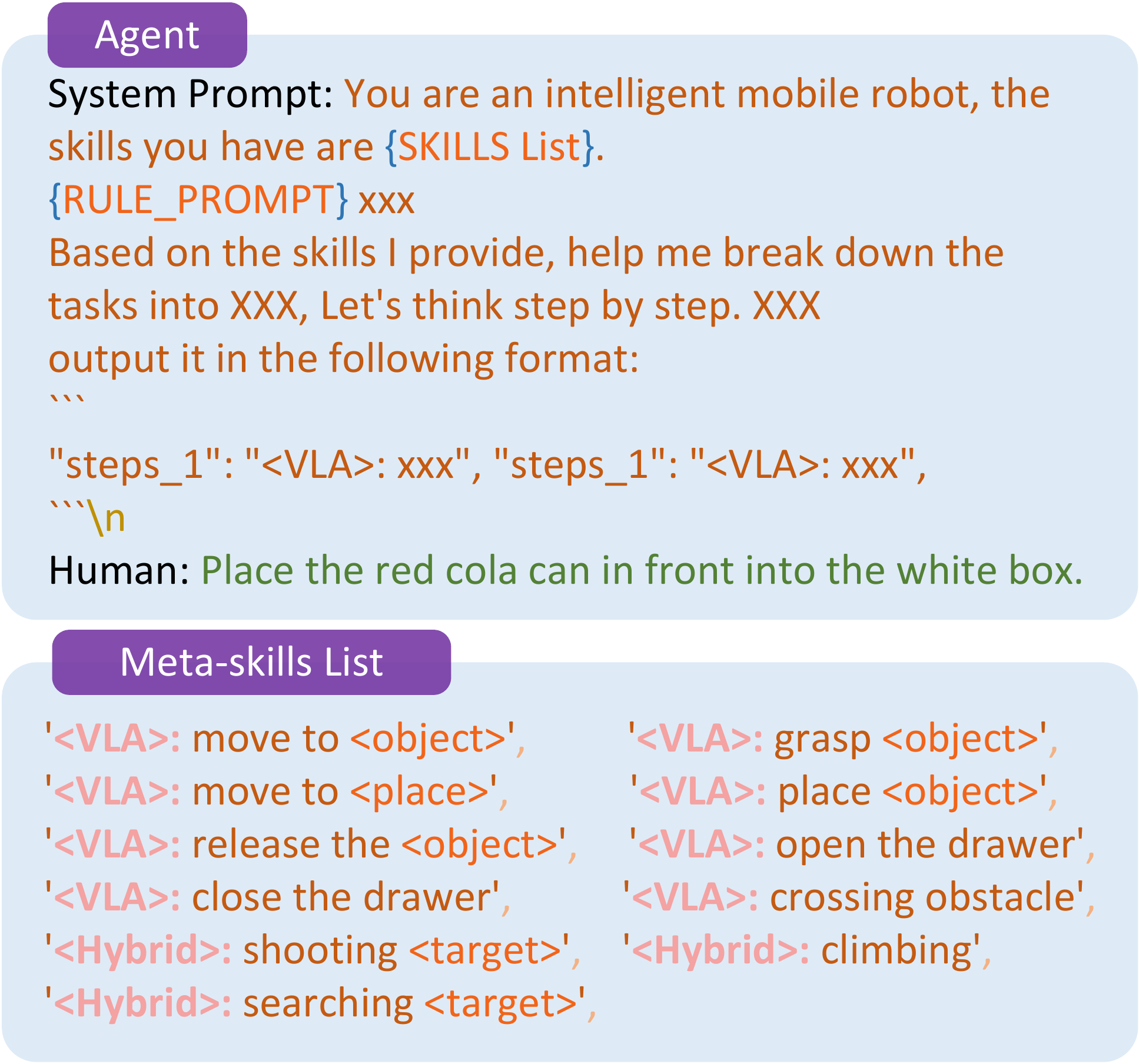}
   \caption{\textbf{The agent prompt and meta-skills list.} }
   \label{fig:prompt}
   \vspace{-10pt}
\end{figure}

\subsection{Skill Models}
\label{sec:skill-models}

For tasks in unstructured environments, such as object manipulation and grasping, the marked generalization ability of LLM-based models allows for handling uncertainties from components, such as object placement, orientation, and category, as well as other unpredictable factors in the environment.
On the other hand, when tasks are executed in specific environments (shooting, searching, and climbing) where the state of the robot and control objectives are of high determinacy, existing traditional models are capable of obtaining superior control performance.
Therefore, we build a more adaptable skill model, including VLA-based and hybrid models, to maximize the performance of each expert model.

\begin{figure*}[!t]
  \centering
   \vspace{-5pt}
   \includegraphics[width=0.95\linewidth]{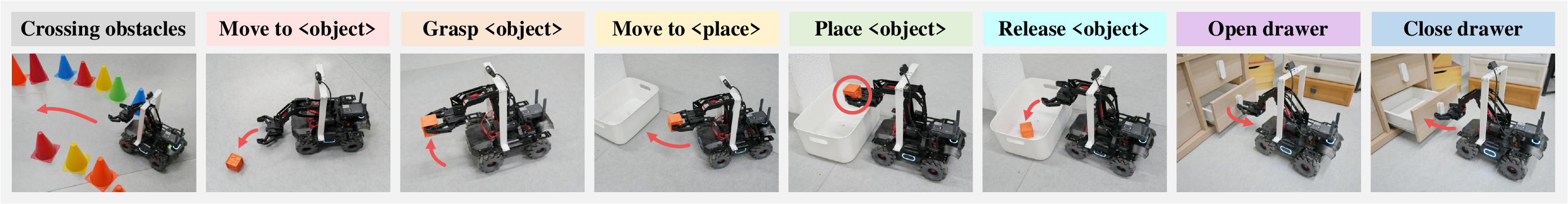}
   \vspace{-5pt}
   \caption{Illustration of meta-skills in the VLA model.}
   \label{fig:eight_skills}
   \vspace{-5pt}
\end{figure*}

\begin{figure*}[!t] 
  \centering
   \includegraphics[width=0.95\linewidth]{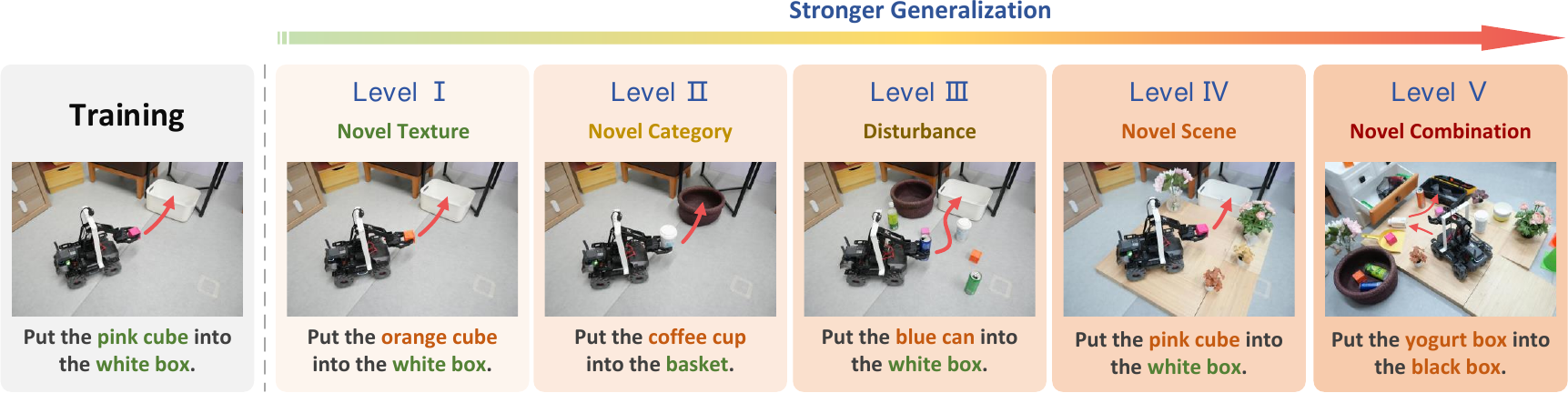}
   \caption{\textbf{Evaluation Protocol for RoboMatrix generalization at the object and scene levels.} Levels \textit{I-II} represent object generalization difficulty, Level \textit{III} serves as a transition, and Levels \textit{IV-V} correspond to scene generalization. Difficulty increases progressively from Level \textit{I} to Level \textit{V}.}
   \label{fig:level}
\end{figure*}

\begin{figure*}[htbp]
  \centering
   \includegraphics[width=0.9\linewidth]{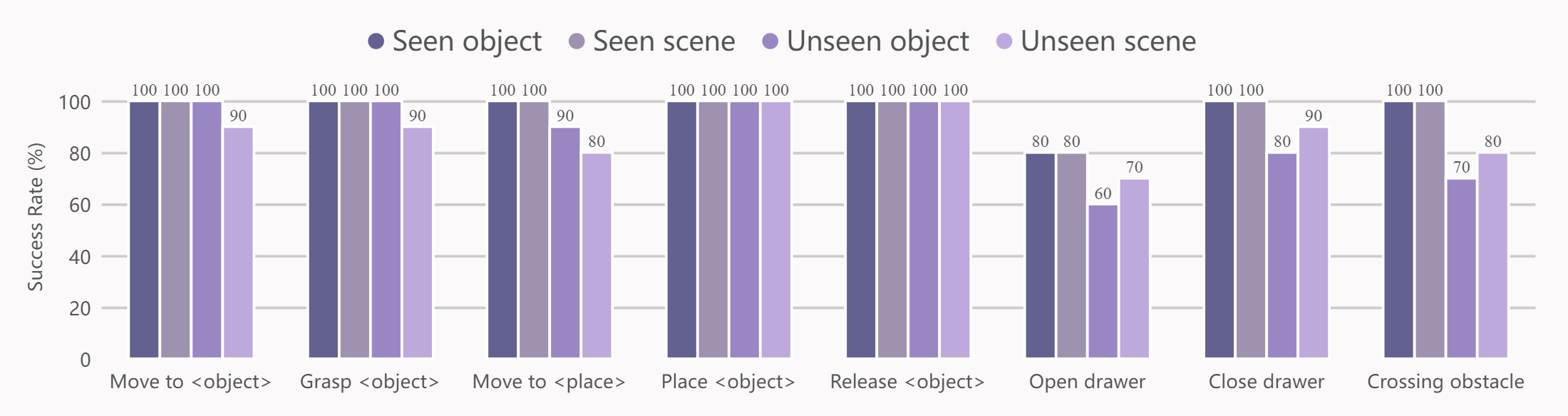}
   \vspace{-5pt}
   \caption{\textbf{Success Rate of meta-skills in the VLA model.} This is the final skill model trained on the full dataset.}
   \label{fig:eight_skill}
   \vspace{-10pt}
\end{figure*}

\subsubsection{Vision-Language-Action Model}
\label{sec:vla}
Our VLA skill model is built upon the decode-only LLM, Vicuna 1.5~\cite{vicuna2}, which is trained based on LLaMA2~\cite{LLAMA2}. The vision encoder uses a CLIP-Large~\cite{CLIP} with an input size of $336\times336$px, followed by two linear layers for visual embedding projection. The entire model takes the images and skill prompts as inputs and produces discrete actions. To maintain higher stability of LLM output, we project the continuous actions into discrete bins following~\cite{RT1, RT2, RT-H}. 
After a comprehensive statistical analysis of the collected multi-robot data, we set the optimal number of discrete bins to 256.
It is worth noting that while RT-2 chooses to overwrite the 256 low-frequency used tokens, we add 256 special tokens to avoid disrupting the original vocabulary.
Our discrete actions are divided into 7 dimensions, with each dimension containing 256 bins, as represented by the following formula:
$$\epsilon, \Delta X, \Delta Y, \Delta \theta_{yaw}, \Delta \mu_{pos}, \Delta \nu_{pos}, \phi$$
where $\epsilon$ represents the stop signal, which is used to determine whether a single skill operation is completed. $\Delta X, \Delta Y, \Delta \theta_{yaw}$ respectively represent the differences in the X-Y position and rotation angle on the real-world ground plane. $\Delta \mu_{pos}$ and $\Delta \nu_{pos}$ is the end-effector pose of the gripper and $\phi$ is the binary status of the gripper. 

\vspace{-10pt}
\paragraph{Alignment Training.}
To achieve multi-modal alignment, we leverage the pre-trained visual embedding projection from LLaVA 1.5~\cite{llava2}. For alignment in the robotic domain, we freeze the vision encoder while unfreezing the projection and LLM. We then perform co-fine-tuning using multi-modal text-image pairs of web data and our rough image-action pair dataset. 


\vspace{-10pt}
\paragraph{Supervised Fine-tuning.}
We utilize approximately 60K visual-action instruction tuning data from the skill database of finely annotated skill data. During model training, we unfreeze all parameters, including the vision encoder.

\subsubsection{Hybrid Model}
The robot invokes the appropriate traditional control strategy for skills while minimizing the error of a single control variable based on its own sensor data, such as using proportional-derivative (PD) control.
For perception tasks in skill, such as object detection, we adopt YOLO-World~\cite{yoloworld} as an open-world detector.
The implementation details of the hybrid model are in the supplementary material (see Sec.~\ref{sec:rationale}).




\subsection{RoboMatrix Framework}
\label{sec:robomatrix-framework}
The hierarchical design of the RoboMatrix aims to extract meta-skills from various complex tasks, schedule skill models to obtain corresponding policies, and control the real-world robots to action. The framework consists of three layers, as shown in the Fig.~\ref{fig:framework}.


The \textbf{Modular Scheduling Layer} includes a \textit{Task-Planning Agent} built upon the Generative Pre-trained Transformer (GPT)~\cite{GPT4} and LangChain~\cite{LangChain}, as well as an \textit{Execution Checker} based on open vocabulary object detector (OVOD)—Grounding DINO v1.5~\cite{ren2024gdino15}.
The task-planning agent decomposes complex tasks into subtask sequences based on a skill list that contains a collection of prompts for various meta-skills (see Fig.~\ref{fig:prompt}). 
If new skills are generated during task decomposition, they will be manually refined and added to the meta-skills list for future reuse.
Before executing a subtask, the execution checker detects the relevant objects involved and ensures that each subtask is executable under the current conditions, thereby enhancing the overall efficiency and success rate of task execution. Once the object is detected in the image, the skill layer will be prompted. If the object is not detected, the process will be interrupted.
The \textbf{Skill Layer} maps the description of subtasks to robot actions, with the action including a stop signal to
determine whether the current subtask is complete. We already detail the implementation of skill model in Sec.~\ref{sec:skill-models}.
The \textbf{Hardware Layer} is based on a distributed system and manages the controller and stage observer of the robot. The supplementary material (see Sec.~\ref{sec:hardware platform} and Sec.~\ref{sec:robomatrix ros}) provides more details on the hardware layer.

%% file: sec/4_exp.tex
\begin{table}[htbp]
\centering 
\resizebox{1.0\linewidth}{!}{ 
\begin{tabular}{l|c|ccccc}
  \specialrule{1pt}{0pt}{1pt}
  \toprule 
  Method & Dataset & $\mathcal{L.}$\textit{I} & $\mathcal{L.}$\textit{II} & $\mathcal{L.}$\textit{III} & $\mathcal{L.}$\textit{IV} & $\mathcal{L.}$\textit{V} \\
 \midrule
 Task-Centric & Mini & 80\% & 30\% & 20\% & 70\% & 0\% \\
 Skill-Centric & Mini & 90\% & 80\% & 60\% & 80\% & 50\% \\
 Skill-Centric & Full & 100\% & 100\% & 90\% & 100\% & 80\% \\
  \bottomrule
  \specialrule{1pt}{1pt}{0pt}
\end{tabular}
} 
\caption{\textbf{Comparison of Task-Centric and Skill-Centric methods.} $\mathcal{L.}$ means level. For the detailed classification of levels, please refer to Fig~\ref{fig:level}.}
\vspace{-.1in}
\label{tab:level}
\end{table}

\section{Experiments}
\begin{table*}[htbp]
\centering 
\resizebox{1.0\linewidth}{!}
{
    \begin{tabular}{l|c|ccccc}
  \specialrule{1pt}{0pt}{1pt}
  \toprule 
  Method & Overall Suc. & Move to cola can & Grasp can & Move to box & Position can over the box & Release \\
 \midrule
 w/o Pretrain & 30\% & 50\% & 80\% & 40\% & 30\% & 90\%
 \\
 w/ Web Pretrain & 80\%	& 90\%	& 100\%	& 100\%	& 80\% & 100\% \\
 \rowcolor[gray]{.9} w/ Robotics Pretrain &  100\% & 100\% & 100\% & 100\% & 100\% & 100\%\\
  \bottomrule
  \specialrule{1pt}{1pt}{0pt}
 \end{tabular}\vspace{-2mm}
}
\caption{\textbf{Success rates for each step in a sequential long-horizon task under different pretraining methods.} "Overall Suc." represents the success rate of completing the entire task, while the five rightmost columns show the success rates of individual steps in order. The results demonstrate that pretraining significantly improves the performance of the skill model.}
\label{tab:pretrain}
\end{table*}

\begin{table*}[htbp]
\centering 
\resizebox{1.0\linewidth}{!}
{
    \begin{tabular}{l|c|ccccc}
  \specialrule{1pt}{0pt}{1pt}
  \toprule 
  Method & Overall Suc. & Move to cola can & Grasp can & Move to box & Position can over the box & Release \\
 \midrule
 ACT* & - & 70\% & 90\% & 40\% & 60\% & 40\% \\
 OpenVLA & 0\% & 10\% & 90\%	& 10\%	& 10\% & 0\% \\
 \rowcolor[gray]{.9} RoboMatrix &  100\% & 100\% & 100\% & 100\% & 100\% & 100\%\\
  \bottomrule
  \specialrule{1pt}{1pt}{0pt}
 \end{tabular}\vspace{-2mm}
}
\caption{\textbf{Success rates of different methods for each step in a sequential long-horizon task.} "Overall Suc." represents the success rate of completing the entire task. * indicates that ACT cannot complete long-horizon tasks and requires a separate model for each step, whereas other methods employ a single unified model.}
\label{tab:diff_method}
\end{table*}


\subsection{Implementation Details}

\noindent \textbf{Robot Configuration.}
We utilize DJI's RoboMaster series robots as the physical platform for RoboMatrix. Robots of different modalities can be connected to a single computer through a specific network communication protocol, allowing RoboMatrix to control multiple robots simultaneously.
We reorganize the open-source API of RoboMaster within the Robot Operating System 2 (ROS2)~\cite{ROS2} framework to enable more flexible distributed control and efficient scheduling of skill models.
The control mode can be switched simply by changing the mapping of the control signal source, enabling both teleoperation via an Xbox controller and autonomous control through a skill model.

\vspace{-10pt}
\paragraph{Dataset and Annotation.}

We extract data for eight skills from approximately 5,000 episodes of high-quality human demonstrations of long-horizon tasks, using a combination of rule-based and manual-based annotation at appropriate proportions.
Fig.~\ref{fig:eight_skills} illustrates the eight meta-skills for our VLA model, each skill can be executed independently or combined to perform long-horizon tasks. 
We ensure the diversity and comprehensiveness of the data for each skill across various dimensions, including object category, appearance, placement, robot initial state, and scene complexity.
The noise from robot state observations in the raw data is filtered to ensure a uniform distribution across all dimensions of the data.
Furthermore, we compiled these 5k episodes into a full dataset. From the full dataset, we selected 200 episodes across 5 different skills to create a mini dataset. Unless otherwise specified, all ablation experiments are conducted on the mini dataset by default.


\vspace{-10pt}
\paragraph{Data Augmentation.}
We apply data augmentation to the stop frames of each skill to ensure the stability of the stop signal output. These stop frames are replicated to achieve an appropriate proportion within the overall skill data. 

\vspace{-10pt}
\paragraph{Training and Inference.}
The training of the VLA skill model uses 8 A100 GPUs with 80GB of memory, and a batch size of 96. During inference, the VLA model operates on a single A100 GPU. To facilitate efficient deployment, we implement a remote VLA inference server that enables real-time action prediction, allowing robots to be controlled without relying on local computational resources. Throughout all training phases, the VLA model is trained with 1 epoch. In addition, for alignment and SFT training, we use a learning rate of 2e-5 and a warmup ratio of 0.01, following the LLaVA-1.5~\cite{llava2} configuration. For more details, please refer to Sec.~\ref{Sup:exp} of the supplementary materials.




\subsection{Performance on Meta-skills}
\label{sec:Overall_Results}

We conduct a comprehensive evaluation of eight meta-skills (see Fig.~\ref{fig:eight_skills}) with the VLA model. Unless otherwise specified, all experiments in this paper are tested with 10 times by default.
As shown in the bar chart in Fig.~\ref{fig:eight_skill}, the results of seen objects and seen scenes demonstrate the strong performance of our skill model. The strong performance on unseen objects and unseen scenes further validates the generalization capability of our skill model. Most skills exhibit slight performance degradation when applied to unseen scenes in comparison to those applied in seen ones. However, for the ``Release \textless object\textgreater'' and ``Place \textless object\textgreater'' skills, our VLA model demonstrate performance levels that are comparable to those counterparts in seen scenes.

\subsection{Performance on Task-level Generalization}

\paragraph{Evaluation Protocol.}
Building on VIMA~\cite{jiang2022vima}, we introduce a 5-level generalization evaluation protocol (see Fig.~\ref{fig:level}. Due to the complexity of evaluation in open-world environments, our metrics primarily evaluate object and scene generalization. Levels \textit{I-II} represent object generalization difficulty; Level \textit{III} serves as a transition, and Levels \textit{IV-V} correspond to scene generalization. Difficulty increases progressively from Level \textit{I} to Level \textit{V}.
Levels \textit{IV-V} primarily assess object generalization, with the distinction between them based on the difficulty of object recognition. Levels \textit{III-V} focus on scene generalization, with their differences primarily determined by the complexity of the scenes.

\begin{figure}[!t]
  \centering
   \includegraphics[width=0.9\linewidth]{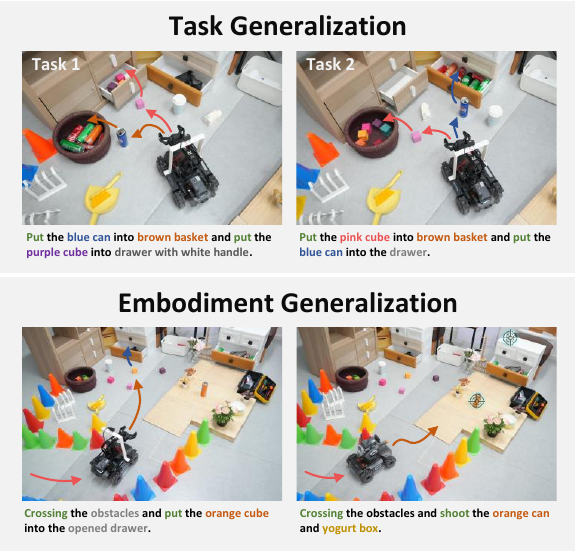}
   \caption{\textbf{Generalization at the task and embodiment levels.} }
   \label{fig:robot_task}
   \vspace{-10pt}
\end{figure}

\vspace{-10pt}
\paragraph{Generalization.}
For \textbf{object} and \textbf{scene} generalization, Tab.~\ref{tab:level} presents the performance comparison between the task-centric and our skill-centric VLA model on the mini and full datasets.
Our method slightly outperforms the task-centric approach in simpler levels, while in more challenging levels, the skill-centric approach significantly outperforms its counterpart. These results show that the skill-centric approach offers clear advantages for difficult and long-horizon tasks.
For \textbf{task} and \textbf{embodiment} generalization, we conduct experiments on two types of long-horizon tasks (see Fig.~\ref{fig:robot_task}), each requiring the execution of ten meta-skills while controlling for the scene and manipulated objects. Additionally, we directly deploy the model trained on the EP robot to the S1 robot for obstacle crossing and shooting tasks. 

\begin{figure*}[!t] 
  \centering
   \includegraphics[width=0.9\linewidth]{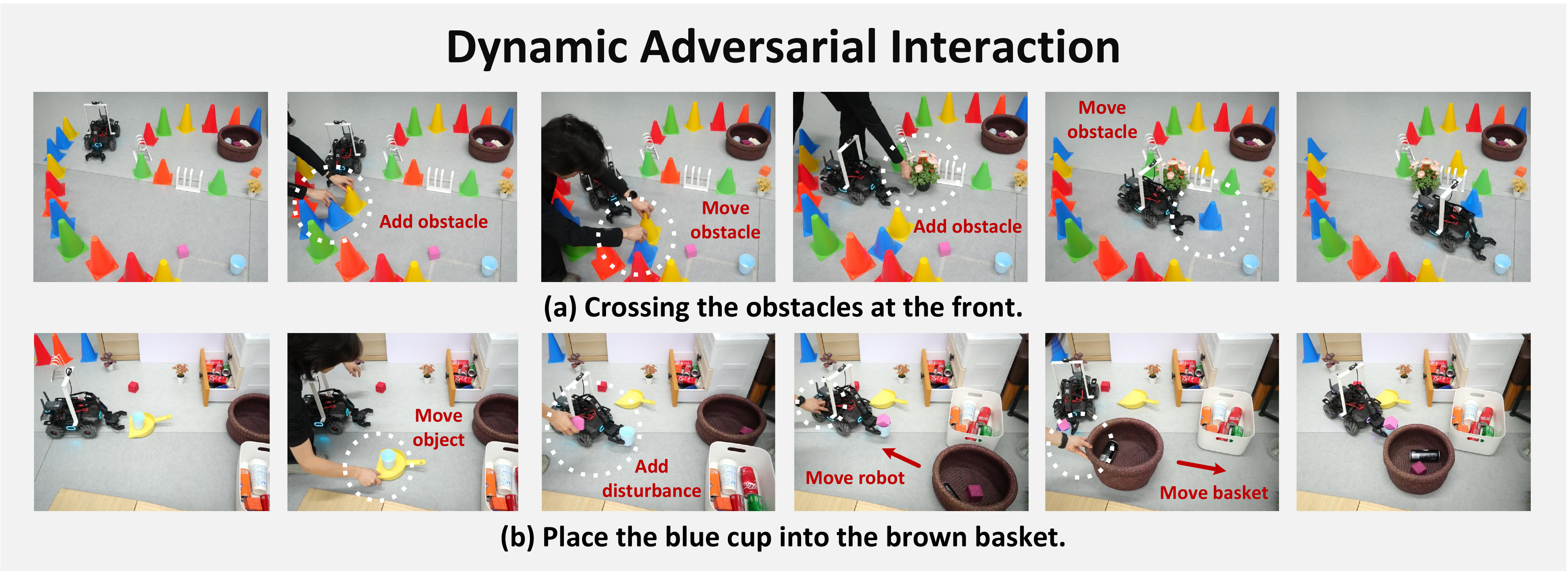}
   \caption{\textbf{Dynamic Adversarial Interaction.}
    Our method demonstrates significant robustness against external dynamic disturbances.}
   \label{fig:interaction}
   \vspace{-10pt}
\end{figure*}

\vspace{-10pt}
\paragraph{Dynamic Adversarial Interaction.} We introduced a substantial variety of unknown human interferences during the execution of various complex tasks. The robustness demonstrated in the experiments (see Fig.~\ref{fig:interaction}) proves the high performance of the skill-centric approach. \textbf{}

\subsection{Ablation Study}
\noindent \textbf{Pretraining.}
In Tab.~\ref{tab:pretrain}, we present three experimental settings designed to demonstrate the necessity and significance of the alignment training discussed in Sec~\ref{sec:vla}. The ``w/o Pretrain'' setting refers to the VLA model with only supervised fine-tuning (SFT) on robot data without any alignment training. The ``w/ web pretrain'' setting involves using the LLaVA-665K~\cite{llava1} dataset for multi-modal alignment. The 'w/ Robotics Pretrain' setting integrates co-fine-tuning with both LLaVA-665K and robot skill data, followed by SFT. The results in the table clearly indicate that multimodal alignment is highly effective, and the alignment within the robot domain further enhances performance. 

\vspace{-10pt}
\paragraph{Mode Size.}
In the field of large language models, increasing model parameters generally means stronger generalization and understanding capabilities. Tab.~\ref{tab:size} demonstrates that this principle holds true for VLA models as well. 
Except for model size, all other experimental settings, including alignment training and supervised fine-tuning (SFT), remain the same across the models. 
The larger 13B model consistently achieves higher success rates across all tasks, especially in unseen scenarios and tasks, which require long-horizon planning.

\vspace{-10pt}
\paragraph{Long-Horizon.}
Tab.~\ref{tab:long_horizon} presents an ablation study on long-horizon tasks with varying difficulty levels. Generally, as the task horizon increases, the difficulty level rises. For easy tasks, the success rates of task-centric and skill-centric methods are comparable. However, for medium long-horizon tasks, the skill-centric approach outperforms the task-centric method by 20\% and this performance gap further widens to 40\% for hard tasks. 
Therefore, the advantage of our skill-centric method becomes more pronounced as the task horizon increases for long-horizon tasks. 

\vspace{-10pt}
\paragraph{Different methods.} As shown in Tab.~\ref{tab:diff_method}, ACT performs well on individual tasks but struggles with multi-task execution. OpenVLA excels at grasping tasks but is less effective in movement-related tasks.


\begin{table}[htbp]
  \centering
  \resizebox{1.0\linewidth}{!}
  {
  \begin{tabular}{c| cc |cc |c}
    \toprule
    \multirow{2}{*}{Size} & \multicolumn{2}{c|}{Moving Suc.} & \multicolumn{2}{c|}{Grasping Suc.} & Long-Horizon \\
    \cmidrule(r){2-6}
    & Seen & Unseen & Seen & Unseen & Unseen \\ 
    \midrule
    7B & 90\% & 70\% & 100\% & 80\% & 70\% \\
    13B & 100\% &100\% & 100\% & 90\% & 100\% \\ 
    \bottomrule
  \end{tabular}
  }
  \caption{\textbf{Ablation study on different model sizes of Vicuna 1.5.} The larger 13B model consistently achieves higher success rates across all tasks, particularly in unseen scenarios and long-horizon tasks.}
  \label{tab:size}
  \vspace{-5pt}
\end{table}



\vspace{-10pt}
\begin{table}[htbp]
  \centering
  \resizebox{1.0\linewidth}{!}
  {
  \begin{tabular}{l|c|ccc}
    \toprule
    Method & Average & Easy & Medium & Hard \\
    \midrule
    Task-Centric & 73\% & 100\% & 80\% & 40\% \\
    Skill-Centric & \textbf{93\%} & 100\% & \textbf{100\%} & \textbf{80\%} \\
    \bottomrule
  \end{tabular}
  }
  \caption{\textbf{Ablation Study on Long-Horizon Tasks with Varying Difficulty.} Easy denotes long-horizon tasks with 3 steps, Medium represents tasks with 5 steps, and Hard includes tasks bigger than 5 steps in unseen scenarios.}
  \label{tab:long_horizon}
  \vspace{-10pt}
\end{table}




%% file: sec/5_conclusion.tex
\section{Conclusion}



In this work, we present a skill-centric hierarchical framework for scalable robot task planning and execution in open-world environments, addressing the need for adaptable and efficient robot control in complex scenarios. A key innovation of our framework is a unified Vision-Language-Action (VLA) model specifically designed for movement and manipulation, which integrates both movement and manipulation outputs to enable versatile robotic actions. Additionally, our framework demonstrates robust generalization across multiple dimensions, including object, scene, task, and multi-robot generalization, underscoring its adaptability and potential for diverse applications. Collectively, these contributions represent a substantial advancement in scalable and generalizable robot autonomy.



%% file: sec/X_suppl.tex
\clearpage
\setcounter{page}{1}
\maketitlesupplementary

\section{Hardware Platform}
\label{sec:hardware platform}
\begin{figure}[!th]
  \centering
   \includegraphics[width=1.0\linewidth]{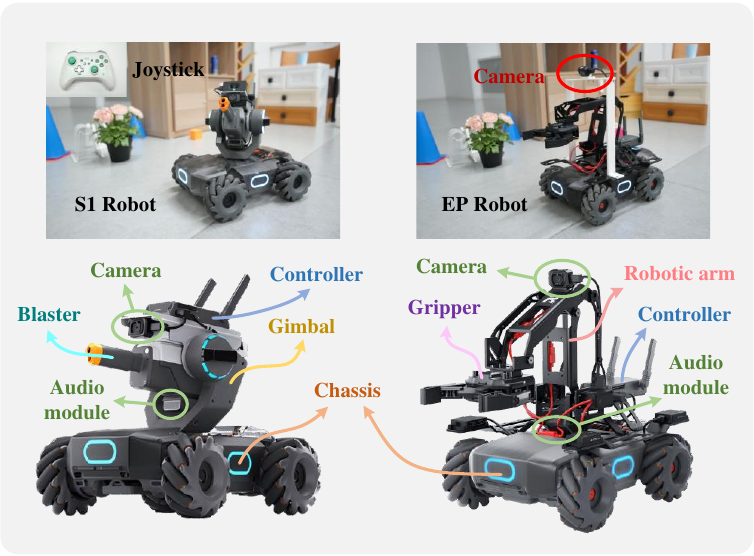}
   \caption{\textbf{RoboMaster platform from DJI.} We modified the EP robot by mounting the camera above the robot to prevent the camera's viewpoint from changing with the movement of the robotic arm. We use a joystick to enable teleoperation control of both the EP robot and the S1 robot.}
   \label{fig:hardware}
   \vspace{-10pt}
\end{figure}

In this section, we introduce the hardware platform of RoboMatrix, as shown in Fig.~\ref{fig:hardware}.

\subsection{RoboMaster Robot}
We use robots from DJI's RoboMaster series as the hardware platform, including the Engineering Robot (EP) and the Warrior Robot (S1).
These two forms of robots share some common components, including the mobile chassis, monocular RGB camera, audio module, and controller.
Additionally, each robot is equipped with a unique set of components to perform specific tasks, such as the target shooting capability of the S1 robot and the target grasping capability of the EP robot.

\vspace{-10pt}
\paragraph{Chassis.}
The mobile chassis is equipped with Mecanum wheels, which provide omnidirectional mobility. This configuration enables decoupled translational movement and rotation in place.
The built-in Inertial Measurement Unit (IMU) allows real-time calculation of the robot's position and orientation relative to a reference coordinate system, with an update frequency of up to 50 Hz.

\vspace{-10pt}
\paragraph{Camera and audio module.}
The monocular RGB camera can capture video streams at a resolution of 1280 $\times$ 720 pixels and 30 FPS. The audio module is capable of capturing environmental audio and playing pre-recorded sound. Notably, we adjust the camera position on the EP robot to stabilize its viewpoint (120$^\circ$). The optimal range for the audio module to receive commands is within 2\ \text{m}.
 
\vspace{-10pt}
\paragraph{Gimbal and Blaster.}
These are components exclusive to the S1 robot. The blaster is mounted on a 2-degree-of-freedom gimbal, allowing rotation along both pitch and yaw angles. Its sight is aligned with the camera, and it is capable of firing bullets with an initial velocity of up to 26\ \text{m/s}.

\vspace{-10pt}
\paragraph{Robotic arm and Gripper.}
These are components exclusive to the EP robot. The gripper is mounted on a 2-degree-of-freedom robotic arm, and due to the unique linkage mechanism design of the arm, the gripper can always remain horizontal.
The forward and inverse kinematics of the robotic arm are easy to compute. The gripper's actions are binary, consisting only of opening and closing.

\vspace{-10pt}
\paragraph{Controller.}
By using a designated application software to connect the controller to the local area network, computers within the same network can control the robot through the official software development kit (SDK), including controlling the robot's various modules and retrieving data from its various sensors.
The delay in control signals depends on the network quality, typically around 100 ms.
Notably, a single computer can scan all the robots within the network and control the robot with a specific serial number.

\subsection{Teleoperation}
We use a joystick for teleoperation of the robot, with the control signals from the joystick mapped to the robot's control system.

\vspace{-10pt}
\paragraph{Robot-independent Module.}
The input from the joystick is mapped to the translational velocity vector of the chassis, with rotational velocity added via the buttons. The target velocity is then calculated into the motor speeds to control the movement of the chassis.

\vspace{-10pt}
\paragraph{Robot-specific Module.}
Whether for the EP robot or the S1 robot, the control of specific modules can be abstracted as the control of a 2-degree-of-freedom mechanism along with an action command for the end-effector.
The input from a set of hat switches is mapped to changes in the robotic arm's end-effector position or the gimbal's orientation. Meanwhile, the input from a single button is mapped to the opening and closing of the gripper, as well as the firing of the blaster.

\section{Hybrid Model}
\label{sec:rationale}

\begin{figure*}[!htbp] 
  \centering
   \includegraphics[width=1.0\linewidth]{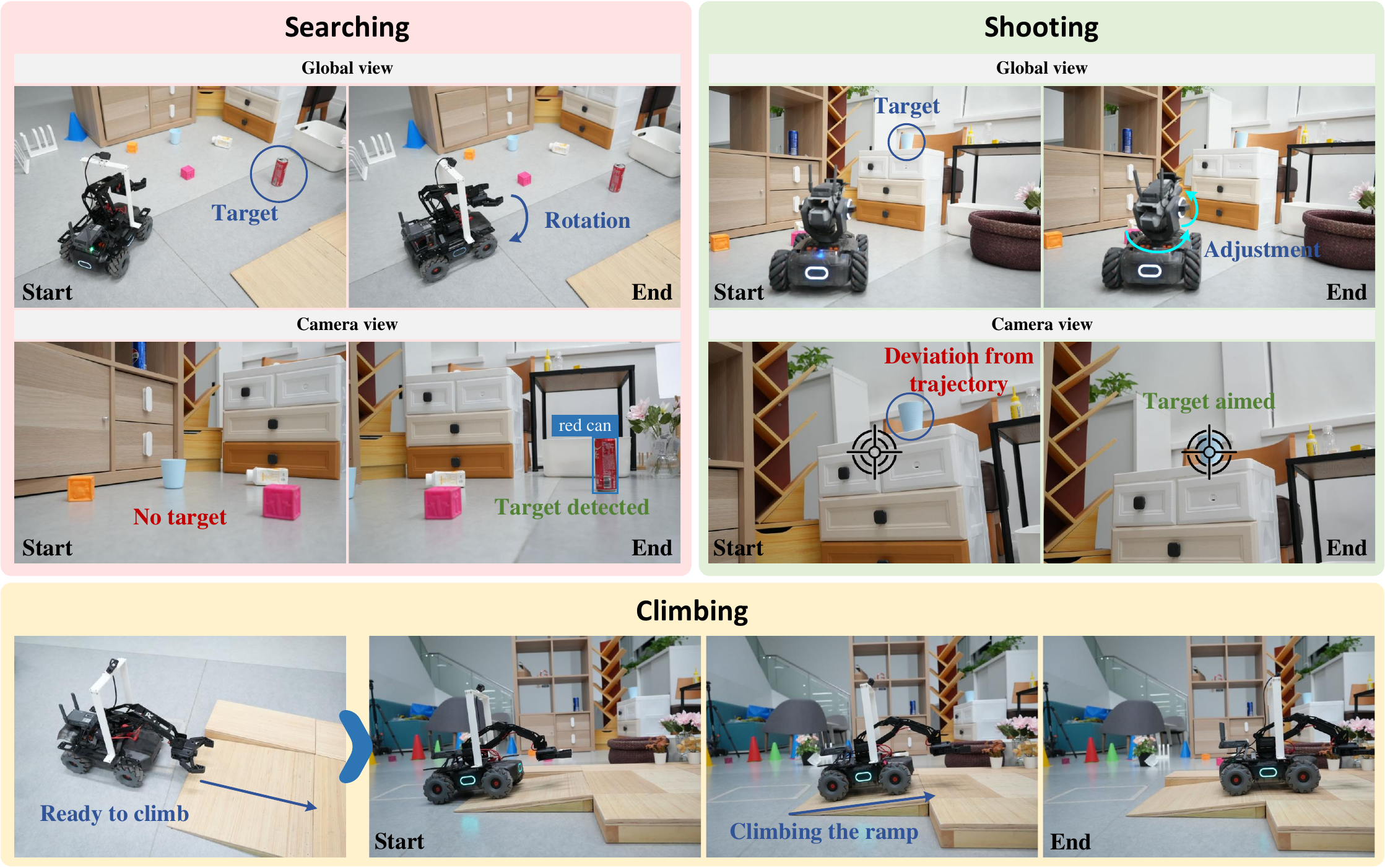}
   \caption{\textbf{Skills in hybrid model.}
   \textbf{Searching:} The robot actively searches for a specific target within the environment with the aim of bringing it into the camera's field of view.
   \textbf{Shooting:} The robot uses a blaster to actively shoot a specific target with the aim of knocking down it.
   \textbf{Climbing:} The robot starts at the bottom of the ramp and actively climbs it with the aim of reaching a raised platform.
   }
   \label{fig:hybrid_model}
   \vspace{-10pt}
\end{figure*}

In this section, we present the implementation details of the hybrid model in RoboMatrix, as shown in Fig.~\ref{fig:hybrid_model}.

\subsection{Searching}
The robot actively searches for a specific target within the environment with the aim of bringing it into the camera's field of view.

Adjusting the camera angle on the EP robot requires changing the position or orientation of the entire chassis because the camera is rigidly attached to the robot. Consequently, controlling the chassis alone is sufficient to modify the camera's viewpoint.
On the other hand, since the camera on the S1 robot is mounted on a gimbal, the viewpoint can be adjusted by controlling the gimbal. By setting the robot’s motion control mode to "gimbal lead," the chassis can follow the gimbal’s movement, enabling synchronized motion between the camera and the robot’s base.

A suitable angular velocity is set to control the rotation of the chassis or the gimbal. As the robot performs a full 360-degree scan, it captures images of the environment at a defined frequency. This image, along with the name of the target, is processed by a lightweight open vocabulary object detector (YOLO-World) to identify whether the specified object is present within the robot's current field of view.
When the robot detects the target, it stops rotating.

\subsection{Shooting}
The robot uses a blaster to actively shoot a specific target with the aim of knocking down it.

Since the crosshair of the blaster is aligned with the center of the camera, it is necessary to control the movement of the gimbal to ensure that the target object is positioned at the center of the camera's field of view.
This process is similar to a visual servo control strategy, where the controller can be built based on Proportional-Derivative (PD) control.

The target's bounding box in the current image is obtained at a certain frequency using the YOLO-World detector. The control signal for the gimbal is calculated based on the relative position between the center of the bounding box and the center of the image, and the robot continues to adjust the gimbal until the positional difference falls within an acceptable tolerance range.
Considering the effect of gravity, the crosshair of the blaster is adjusted based on the distance information from sensors (infrared distance sensor), slightly above the target object.

\subsection{Climbing}
The robot starts at the bottom of the ramp and actively climbs it with the aim of reaching a raised platform.

Under the condition that the robot's chassis is aligned with the ramp, a reasonable speed value is assigned based on the ramp's gradient to control the robot's movement up the ramp and prevent it from sliding down.
The ramp's gradient can be calculated using the robot's built-in sensors (Inertial Measurement Unit), which corresponds to the robot's attitude (pitch angle). The robot is commanded to stop moving when it reaches the platform, as indicated by a pitch angle of zero.

\begin{figure}[!b]
  \centering
   \includegraphics[width=1.0\linewidth]{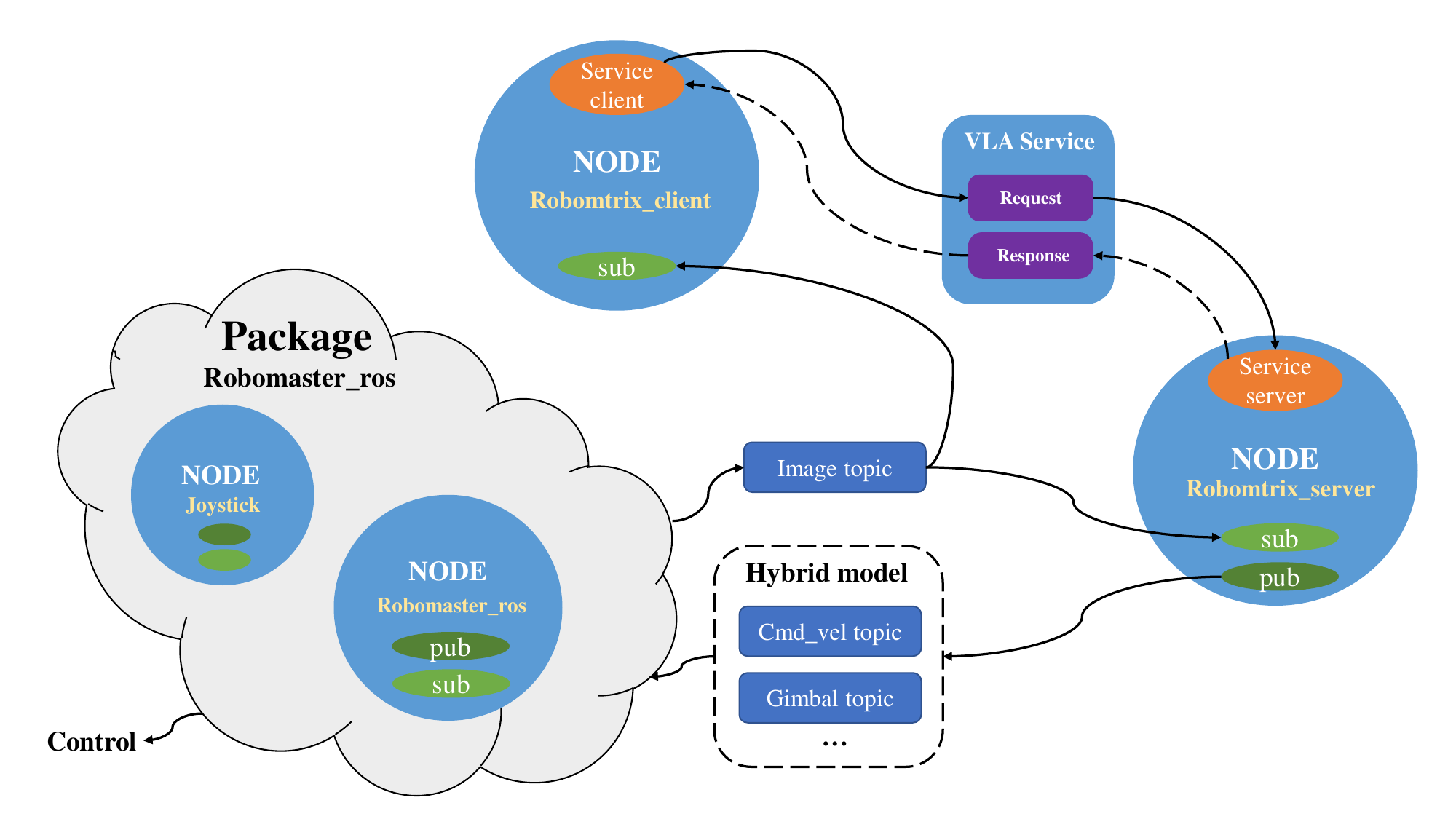}
   \caption{\textbf{Node Graph of Robomatrix System.} }
   \label{fig:robo_ros}
\end{figure}

\section{Additional Experiment Details}

\subsection{More Details on Experiment Setting}
\label{Sup:exp}

\paragraph{Training.}
We conduct alignment training for approximately 180 hours, utilizing 8 A100 GPUs. The pretraining data is broader but lower in quality, helping the model learn various strategies and recover from mistakes. During the supervised fine-tuning (SFT) stage, we train for approximately 30 hours under the same setting. The SFT data is more focused, using high-quality human-annotated data to teach the model how to complete tasks through a skill-centric strategy.



\subsection{More Details on Dataset}


\paragraph{Human Annotation.}
To acquire high-quality, skill-centric data for the supervised fine-tuning (SFT) stage, we employ many annotators to label those data. Although these annotators initially lack relevant experience, they quickly develop the necessary annotation skills through expert-led training. For the collected skill videos, the annotators remove invalid segments from the beginning and end, discarding entire segments of poorly executed data. Additionally, they assign a specific skill name to each valid skill video.

\vspace{-10pt}
\paragraph{Absolute vs. Relative position.}
Regarding the data encoding method, we experiment with two approaches: absolute position and relative position. We discover that with absolute position encoding, the robot struggles to execute tasks successfully, and the model tends to overfit the data, losing its generalization ability. Therefore, we adopt the relative position approach for all our data.

\vspace{-10pt}
\paragraph{Interval Prediction.}

In real-world experiments, we find that when using the current frame image as input and the current frame action as supervision, the trained model predicts actions with small variations, resulting in slow robot motion. We hypothesize that this may be related to the small magnitude of action changes predicted by the model. We experiment with using the current frame image as input and future frame actions as supervision. Ultimately, we discover that using actions from 10 frames ahead for SFT yields the best robot motion performance, ensuring that the robot neither moves too slowly nor too quickly, which could lead to imprecise operations.
Using future frame actions enables the model to learn more forward-looking planning and decision-making capabilities, thereby smoothing and improving the robot's movements.

\begin{figure*}[!t]
  \centering
   \includegraphics[width=1.0\linewidth]{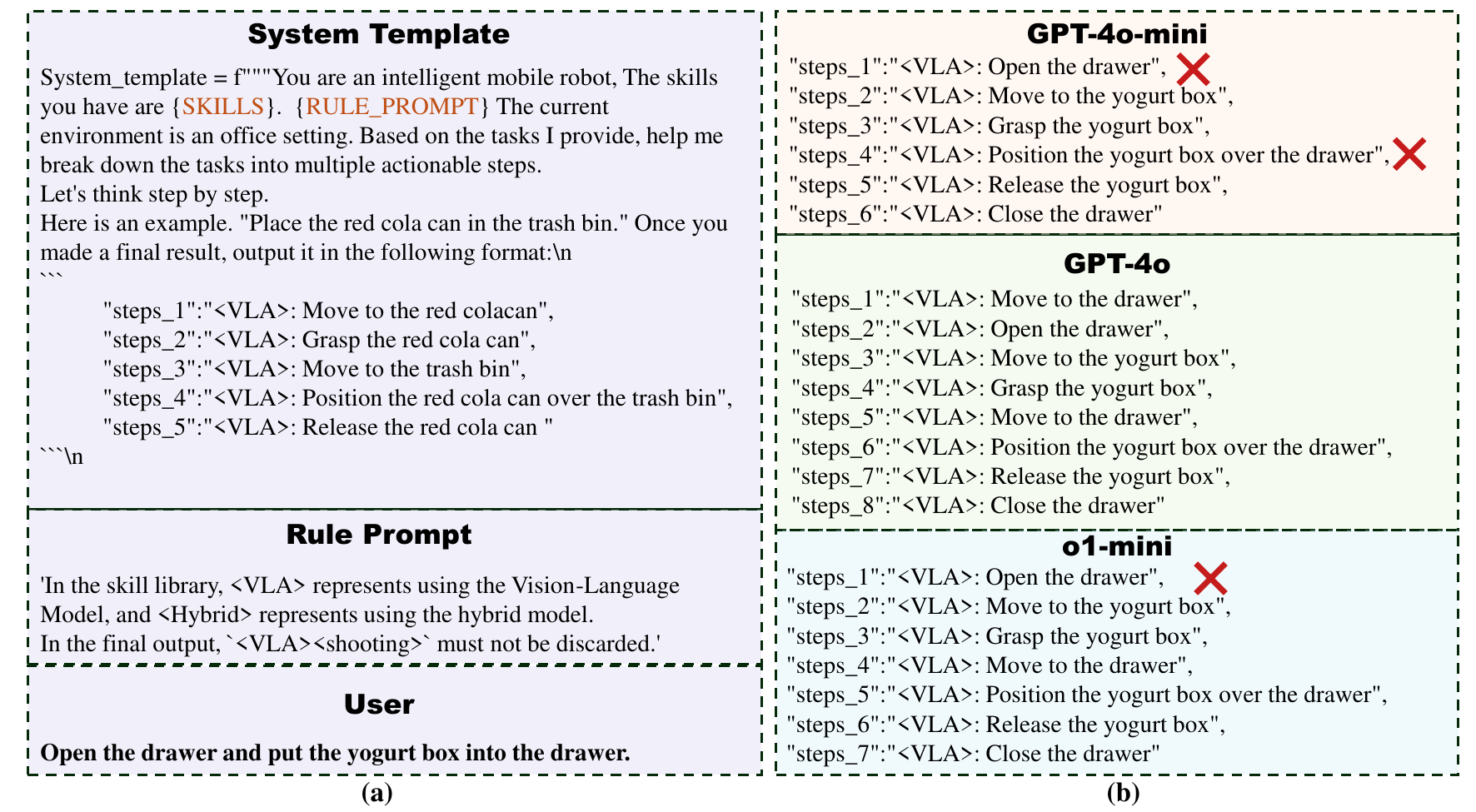}
   \caption{\textbf{Ablation study for different GPT.} (a) shows the prompt used by the planning agent, and (b) shows the output results from different GPT-based agents.}
   \label{fig:supp_prompt}
   \vspace{-10pt}
\end{figure*}

\subsection{More Details on RoboMatrix-ROS}
\label{sec:robomatrix ros}

The entire system is managed using the ROS framework to achieve more modular and efficient communication and control. It is divided into four nodes, as illustrated in Fig. \ref{fig:robo_ros}. The $robomaster\_ros$ package includes both the basic control node and the teleoperation node. It publishes sensor topics and receives control topics for the chassis and gimbal to implement the VLA model or hybrid model. Task planning and management within the system are executed using the ROS service mechanism. The $robomatrix\_client$ node is responsible for task planning and invoking specific VLA skills. Detailed tasks and prompts are sent using custom requests. The implementation of VLA skills is carried out within the $robomatrix\_server$ node, which receives skill names and commands, executes sub-tasks, and returns the execution results. The $robomatrix\_client$ node then receives these results and either sends the next sub-task or proceeds to the planning and management of the next task.

\section{More Experiments}

As shown in Fig.~\ref{fig:supp_prompt} (a), the prompt used in the planning agent includes a sample task description under "User." Fig.~\ref{fig:supp_prompt} (b) presents the output results when the agent uses different GPT models as the foundational model. The figure shows that GPT-4o-mini and o1-mini skipped the step "Move to the drawer" and directly performed "Open the drawer." In this task decomposition scenario, the performance of o1-mini is better than GPT-4o-mini but inferior to GPT-4o.
Overall, the experimental results of GPT-4o align more closely with our expectations.


\section{Additional Visualizations}

\subsection{Assets}

\begin{figure}[!tb]
  \centering
   \includegraphics[width=1.0\linewidth]{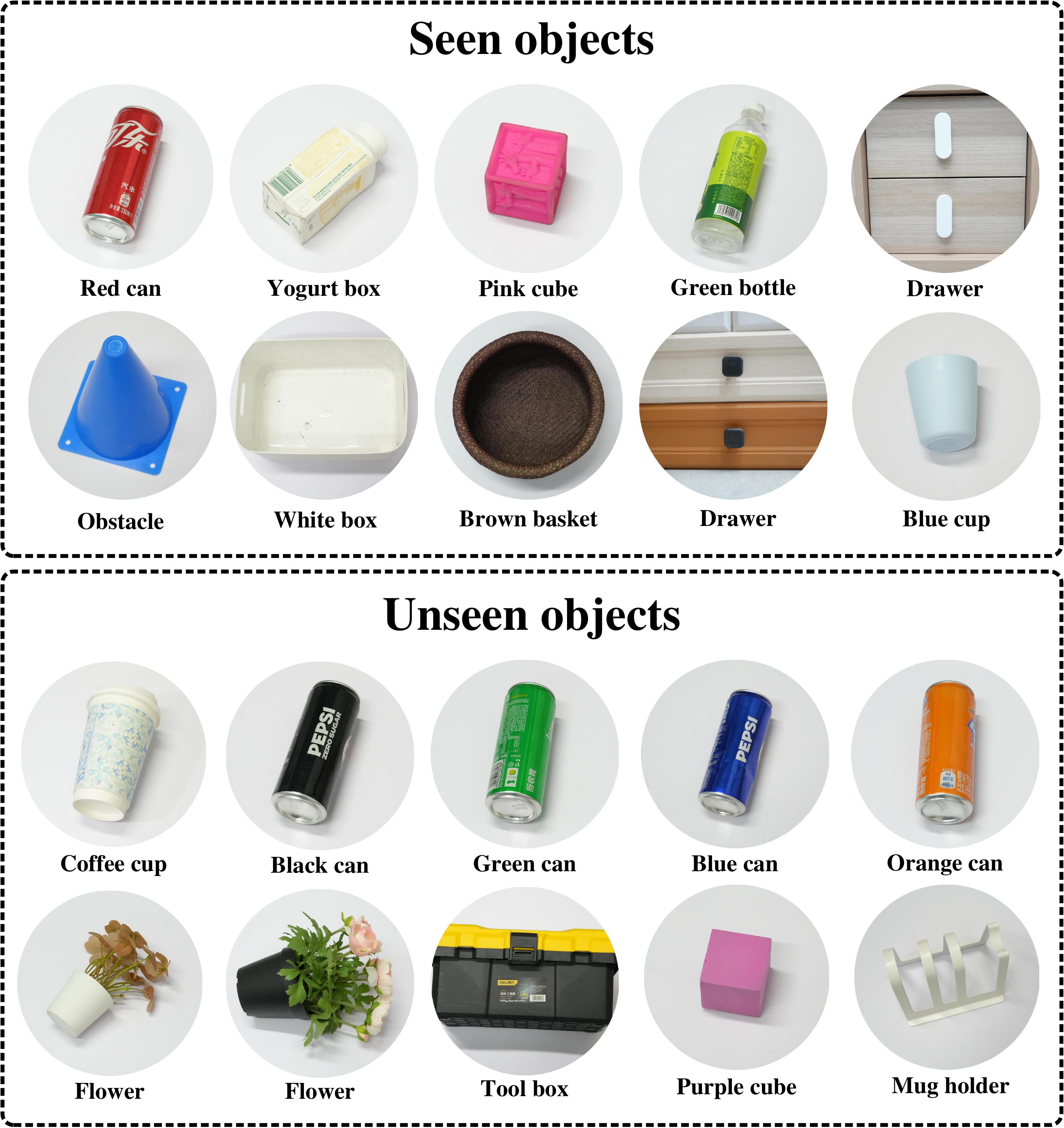}
   \caption{\textbf{Seen objects and unseen objects.} }
   \label{fig:seen_unseen}
   \vspace{-10pt}
\end{figure}

\paragraph{Objects.}
Fig.~\ref{fig:seen_unseen} shows the seen objects used during data collection and the unseen objects during the experiment.

\vspace{-10pt}
\paragraph{Scenes.}
During the data collection process, only relevant objects and a small number of distractors were added to the scene. In the experiment, we created unseen scenarios by altering the types, quantities, and relative positions of objects within the scene.

\subsection{Long-horizon Tasks}

\begin{figure*}[!th] 
  \centering
   \includegraphics[width=1.0\linewidth]{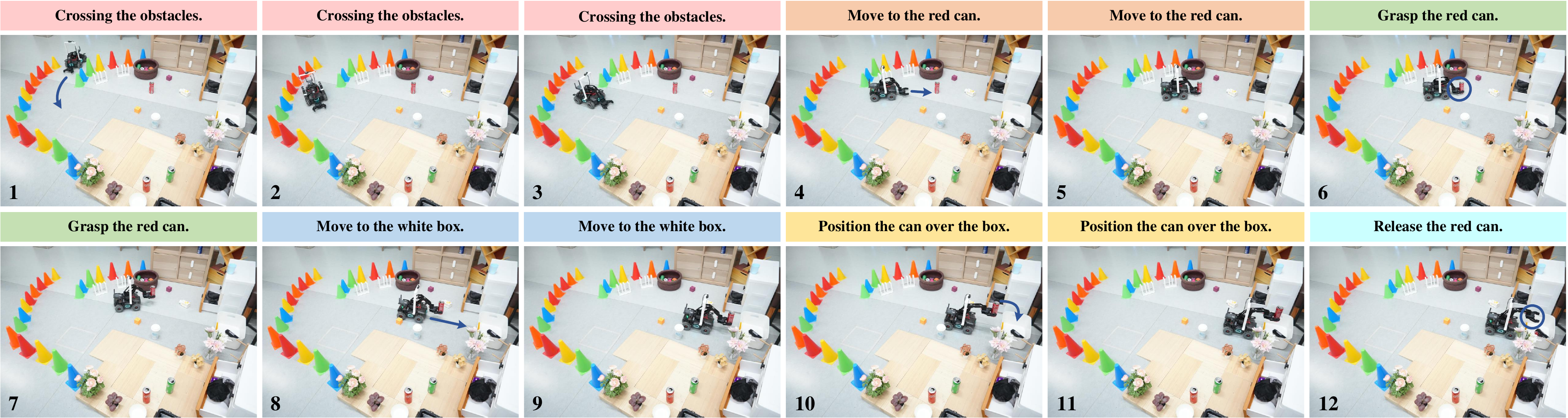}
   \caption{\textbf{Long-horizon task 1:}
   Cross the obstacles at the front and put the red can into the white box.
   }
   \label{fig:move_pick_place_red_can}
\end{figure*}

\begin{figure*}[!th] 
  \centering
   \includegraphics[width=1.0\linewidth]{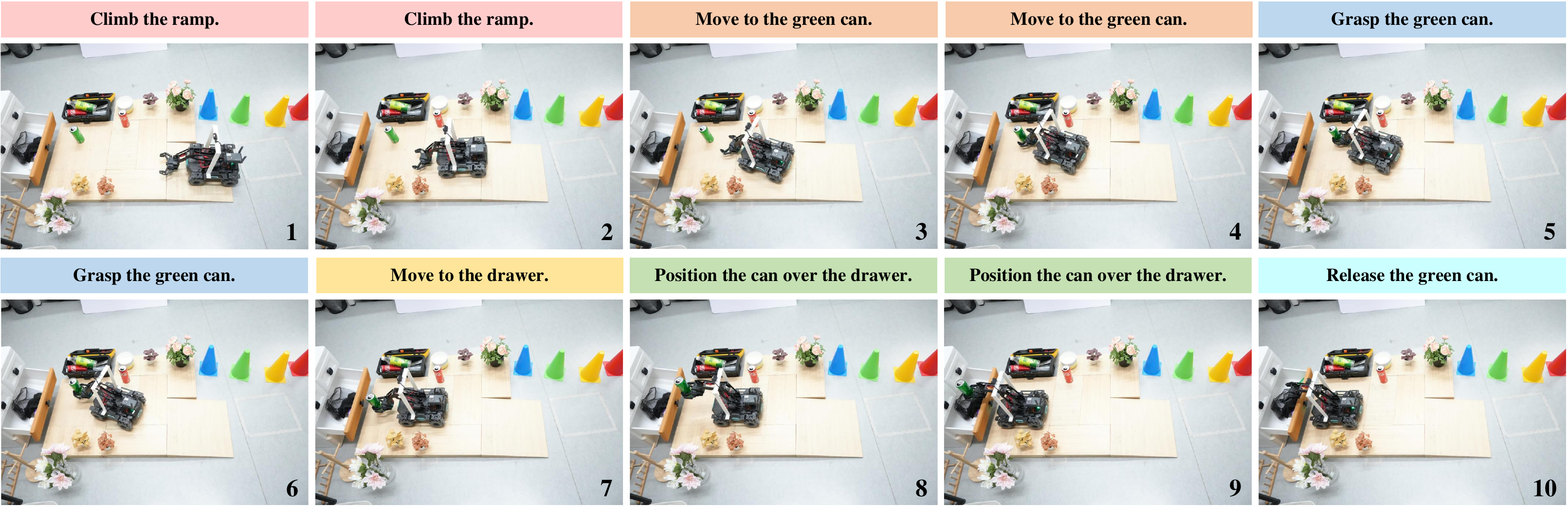}
   \caption{\textbf{Long-horizon task 2:}
   Climb the ramp and put the green can into the drawer.
   }
   \label{fig:pick_drawer}
\end{figure*}

\begin{figure*}[!th] 
  \centering
   \includegraphics[width=1.0\linewidth]{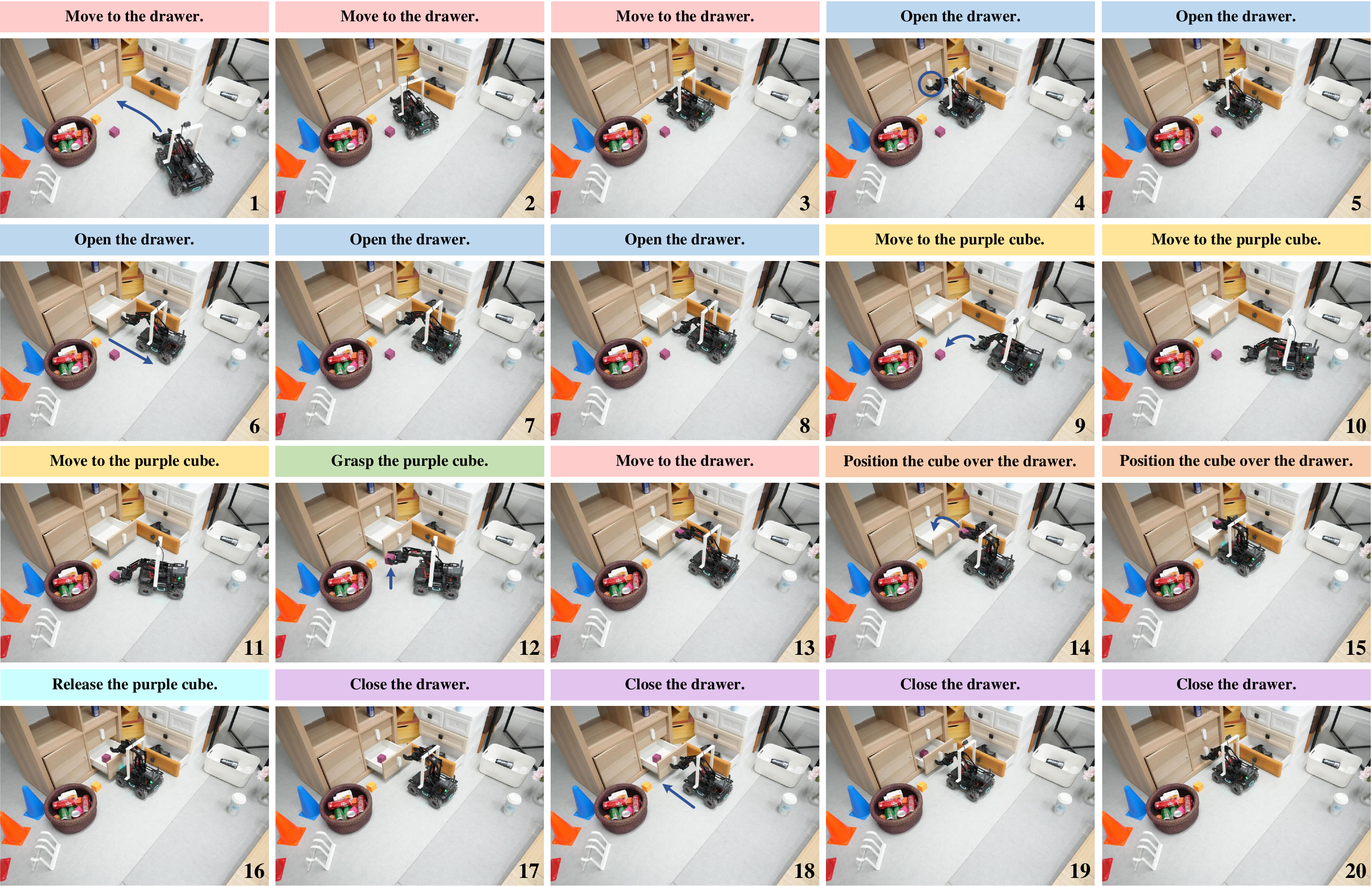}
   \caption{\textbf{Long-horizon task 3:}
   Open the drawer and put the purple cube into the drawer, then close the drawer.
   }
   \label{fig:open_close_drawer}
\end{figure*}

The skill-centric RoboMatrix can exhibit a significant advantage over task-centric approaches in long-horizon tasks.
It can accomplish tasks by reusing existing skills without the need to collect large amounts of additional data.
We validated the capabilities of RoboMatrix on three manually designed long-horizon tasks, demonstrating four levels of generalization as we mentioned in the paper.

\paragraph{Task 1: Cross the obstacles at the front and put the red can into the white box.}
As shown in Fig.~\ref{fig:move_pick_place_red_can}, the EP robot is required to first navigate through obstacles to reach the main scene, then approach and grasp the red can. Finally, it must transport the red can to the white box and place it inside.
Even with changes to the obstacles, the addition of distractions in the scene, modifications to the objects to be grasped, or alterations in their positions, the robot can still successfully complete the task.

\paragraph{Task 2: Climb the ramp and put the green can into the drawer.}
As shown in Fig.~\ref{fig:pick_drawer}, the EP robot first climbs a ramp to reach a platform, then picks up the green can, and finally places it into an open drawer.
It is worth noting that the potted plants in the scene do not interfere with task execution. Additionally, the robot can successfully complete the task even when required to place objects into an unseen black toolbox.

\paragraph{Task 3: Open the drawer and put the purple cube into the drawer, then close the drawer.}
As shown in Fig.~\ref{fig:open_close_drawer}, the EP robot first opens the closed drawer in the scene, then places the purple block next to the drawer inside, and finally closes the drawer.
Even with distractions added next to the drawer, the robot can still complete the task without any interference.